\documentclass[letterpaper, 10 pt, conference]{ieeeconf}  

\IEEEoverridecommandlockouts                              

\usepackage{graphics} 
\usepackage{epsfig} 
\usepackage{times} 
\usepackage{amsmath} 
\usepackage{amssymb}  
\usepackage{graphicx}
\usepackage{colortbl}
\usepackage{xcolor}
\usepackage{url}
\usepackage{hyperref}
\title{\LARGE \bf
VLM Can Be a Good Assistant: Enhancing Embodied Visual Tracking with Self-Improving Vision-Language Models}

\author{Kui Wu$^{1}$, Shuhang Xu$^{2}$, Hao Chen$^{3}$, Churan Wang$^{4}$,  Zhoujun Li$^{1}$, Yizhou Wang$^{4}$, Fangwei Zhong$^{2}$
\thanks{*This work was supported by the National Science and Technology Major Project (2022ZD0114904), NSFC-6247070125, NSFC-62406010, and the Fundamental Research Funds for the Central Universities.}
\thanks{$^{1}$Kui Wu and Zhoujun Li are with State Key Laboratory of Complex \& Critical Software Environment, Beihang University, Beijing, China}
\thanks{$^{2}$ Shuhang Xu and Fangwei Zhong are with the School of Artificial Intelligence, Beijing Normal University, Beijing, China.   Correspondence to {\tt\small fangweizhong@bnu.edu.cn}
}
\thanks{$^{3}$Hao Chen is with City University of Macau, Macao, China}
\thanks{$^{4}$ Churan Wang and Yizhou Wang are with Center on Frontiers of Computing Studies, School of Computer Science, Nat'l Eng. Research Center of Visual Technology, Peking University, Beijing, China.
 }
}

\begin{document}

\maketitle
\thispagestyle{empty}
\pagestyle{empty}

\begin{abstract}
We introduce a novel self-improving framework that enhances Embodied Visual Tracking (EVT) with Vision-Language Models (VLMs) to address the limitations of current active visual tracking systems in recovering from tracking failure. Our approach combines the off-the-shelf active tracking methods with VLMs' reasoning capabilities, deploying a fast visual policy for normal tracking and activating VLM reasoning only upon failure detection. The framework features a memory-augmented self-reflection mechanism that enables the VLM to progressively improve by learning from past experiences, effectively addressing VLMs' limitations in 3D spatial reasoning. Experimental results demonstrate significant performance improvements, with our framework boosting success rates by $72\%$ with state-of-the-art RL-based approaches and $220\%$ with PID-based methods in challenging environments. This work represents the first integration of VLM-based reasoning to assist EVT agents in proactive failure recovery, offering substantial advances for real-world robotic applications that require continuous target monitoring in dynamic, unstructured environments. Project website: \url{https://sites.google.com/view/evt-recovery-assistant}.

\end{abstract}


\section{INTRODUCTION}

Embodied Visual Tracking (EVT) is a critical task for embodied AI, requiring agents to track dynamic targets while navigating through unstructured environments. Unlike traditional visual tracking tasks~\cite{VOT_TPAMI}, EVT requires agents to not only understand their surroundings but also to control their movements and camera angles to continuously monitor a target in an ever-changing context. This capability forms the foundation for numerous real-world robotic applications, such as social navigation and person-following robots~\cite{ci2023proactive,wang2018accurate,jin2022conquering}, which must maintain awareness of a target human in dynamic environments, assistive robots that shadow users while avoiding obstacles.

While recent advancements in EVT, primarily focused on reinforcement learning (RL) methods~\cite{devo2021enhancing, zhou2023deep, dionigi2022vat,dionigi2024d}, have achieved significant progress in maintaining stable tracking in dynamic scenarios, many existing models still face notable limitations. These models typically rely on temporal information and learned policies to handle short-term occlusions or brief losses of target visibility. However, they struggle with more complex situations, such as completely losing track of the target for extended periods, where existing strategies often prematurely declare failure. Furthermore, these methods lack the ability to proactively search for and recover targets after prolonged loss, severely limiting their applicability in complex real-world environments.

Concurrently, the field has witnessed remarkable advances in Vision-Language Models (VLMs)~\cite{zhang2024vision}, which have demonstrated exceptional cross-domain generalization capabilities and powerful scene-reasoning abilities. However, these models have notable limitations in robotics applications. Most VLMs are primarily trained on static images or passive video understanding without action capabilities, resulting in limited spatial reasoning~\cite{chen2024spatialvlm} and navigation planning~\cite{zhang2024navid} in complex 3D environments. Despite these constraints, recent VLMs like GPT-4V have shown impressive zero-shot performance and in-context learning ability in understanding complex visual scenes, which presents a promising direction for addressing the limitations of current EVT approaches.

\begin{figure}[t]
    \centering
    \includegraphics[width=1\linewidth]{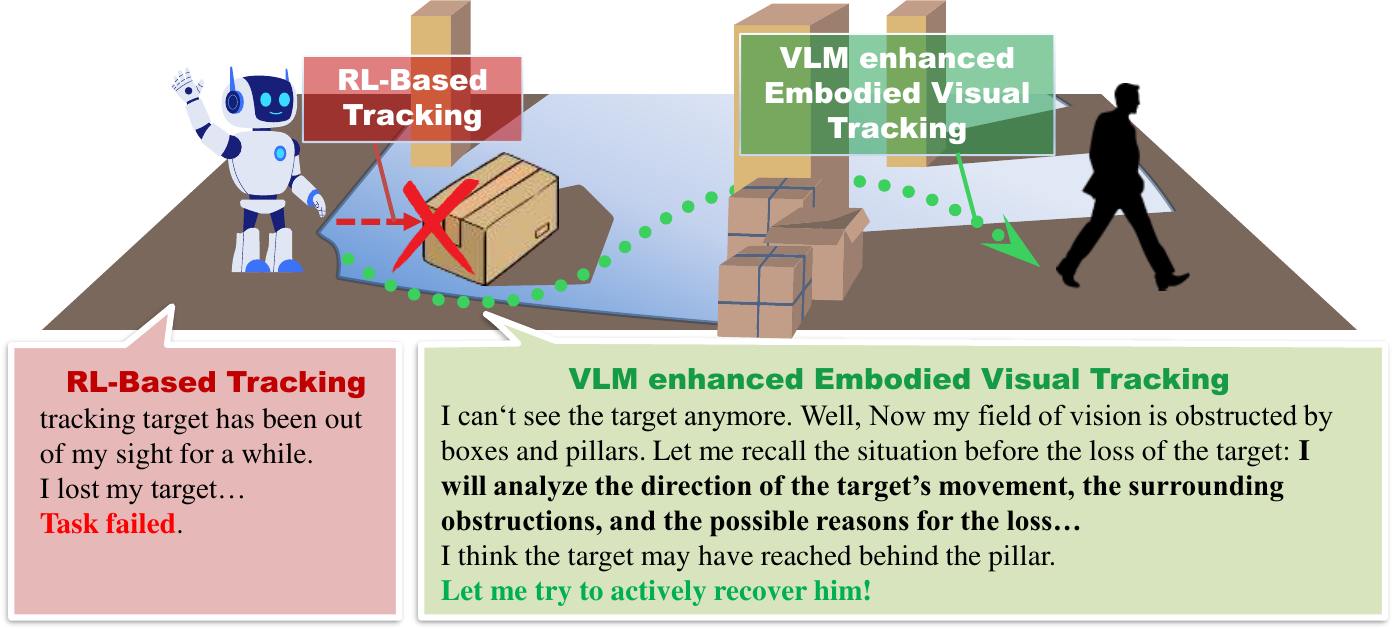}
    \caption{Comparison of tracking capabilities between traditional tracking method and our proposed VLM-enhanced Embodied Visual Tracking approach. When the target is occluded by obstacles (boxes and pillars), the traditional method may lose track and fail (\textcolor[RGB]{168,0,0}{Red}), while our VLM-enhanced approach analyzes the target's trajectory, and surrounding environment, and actively attempts to recover the target by reasoning about its possible location behind the pillar (\textcolor[RGB]{0,168,0}{Green}).}
    \label{fig:example}
\end{figure}

To bridge this gap, we propose a novel self-improving framework that integrates a Vision-Language Model (VLM) to assist EVT agents in recovering from tracking failures. Our approach combines the off-the-shelf traditional tracking algorithms with the powerful reasoning capabilities of VLMs, creating a complementary system that leverages each component's strengths while mitigating their individual weaknesses. Specifically, we employ an RL-based visual tracking policy~\cite{zhong2024empowering} to follow the target in nominal conditions and only activate the more computationally intensive VLM reasoning when failures are detected. The system identifies target loss through a reliable segmentation-based detection mechanism that continuously monitors the target's presence. Upon failure detection, the VLM analyzes the historical observation sequence leading to the failure, interprets the environmental context, and generates a structured sequence of recovery actions as movement suggestions. 
Most importantly, to overcome the inherent limitations of VLMs in 3D spatial reasoning and to optimize recovery strategies for unknown failure cases, we introduce a memory-augmented self-reflection mechanism that enables the VLM to progressively self-improve by analyzing, storing, and learning from its historical experiences. This approach creates a continuously evolving recovery capability that becomes increasingly effective at handling complex tracking failures over time.

Our contributions can be summarized in four-fold:
1) We are the first to introduce a VLM-based reasoner to assist the embodied visual tracking agents to proactively recover from failure cases. 
2) We propose a novel memory-augmented self-reflection mechanism that enables agents to improve through accumulated experience, effectively addressing VLMs' inherent limitations in 3D spatial reasoning.
3) Our VLM-enhanced framework significantly boosts tracking performance across different base policies, increasing success rates by $72\%$ with state-of-the-art RL-based approaches and $220\%$ with PID-based methods in challenging environments.
4) We also conduct comprehensive experiments to analyze the effectiveness of each module in our proposed framework, providing insights into the complementary nature of our memory management and reflection mechanisms and their collective impact on performance.

\section{Related works}

\subsection{Embodied Visual Tracking in Complex Environments}
Visual tracking is a fundamental capability for mobile robots operating in dynamic environments, requiring systems to actively control camera views and maintain awareness of surroundings while following targets~\cite{devo2021enhancing, dionigi2022vat}. Previous research in this domain has approached the challenge from several perspectives. Traditional approaches relied primarily on geometric methods~\cite{choi2012robust, meger2008curious} and filter-based tracking techniques~\cite{schulz2001tracking} to maintain target awareness. While effective in controlled environments, these methods often struggled with occlusions and dynamic scenes. Building on these foundations, specialized person-following robots~\cite{kulkarni2022person, nikdel2018hands} emerged, predictive modeling to maintain line-of-sight tracking. More recently, reinforcement learning (RL) based approaches have made substantial advances in tracking robustness~\cite{luo2018end, zhou2023deep}. Researchers further employ spatial-temporal neural networks and adversarial mechanisms to handle short-term occlusions and distractions ~\cite{zhong2021distractor,zhong2018advat}. Advanced state representation and RL strategies ~\cite{zhong2024empowering} have further improved training efficiency and enabled short-term trajectory prediction for brief occlusion handling. Despite these improvements, existing robotic systems still struggle to proactively recover from significant tracking failures, particularly in scenarios involving complex environmental structures. In this work, we propose a self-improving framework that integrates a Vision-Language Model (VLM) to assist visual tracking agents in recovering from such failures. 

\subsection{Vision-Language Model for Embodied Vision}
Vision-Language Models (VLMs)~\cite{zhang2024vision} have advanced embodied AI through their extensive domain knowledge and multi-modal understanding. 
Researchers have leveraged VLMs in several ways to enhance robots. Zero-shot approaches~\cite{bai2023qwen,liu2023llava,huang2024embodied} apply VLMs directly to tasks like visual question answering (VQA), utilizing the intrinsic image-text relations for more detailed caption. However, VLMs often lack accurate spatial reasoning abilities ~\cite{chen2024spatialvlm}, which limits their effectiveness in tasks requiring precise 3D awareness. To address this, researchers have explored several ways to integrate VLM into the robotic system. For instance, Navid~\cite{zhang2024navid} jointly fine-tuned visual and textual encoders to improve vision-language navigation (VLN). In complex scenarios, VLMs serve as submodules in multi-modal systems. VLFM~\cite{yokoyama2024vlfm} combines BILP-2~\cite{li2023blip} for object recognition with depth and odometry data to generate 2D maps for navigation. NavGPT-2~\cite{zhou2025navgpt} employs GPT-4V with image sequences for stepwise navigation reasoning, adjusting routes dynamically based on visual instructions to enhance indoor navigation. In our work, we building a self-improving reasoning framework to dynamically optimize the VLM's decision-making process to assist the off-the-shelf tracking policy. 
While OpenVLA~\cite{kim2024openvla} introduced end-to-end action control for diverse embodied tasks, our experiments show it accumulates errors in dynamic environments during embodied visual tracking. 
In contrast, our work uniquely builds a self-improving reasoning framework that dynamically optimizes VLM decision-making to assist existing tracking policies, specifically addressing the challenges of prolonged target recovery in visual tracking scenarios.

\begin{figure*}
    \centering
    \includegraphics[width=0.9\linewidth]{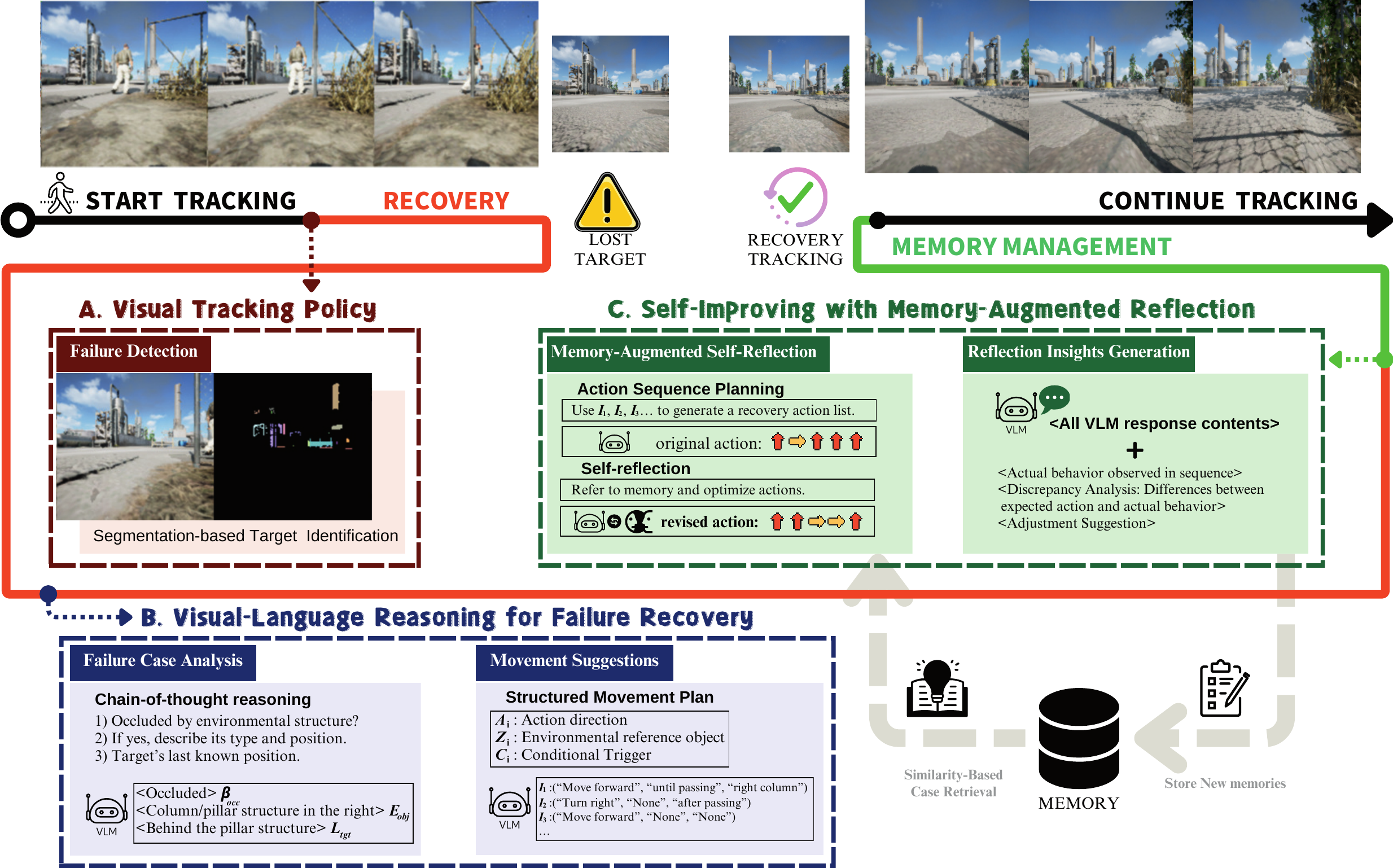}
    \caption{The framework for integrating Vision-Language Models (VLMs) with active tracking policies. The framework follows a structured recovery approach when target tracking fails, consisting of five main steps: (1) \textbf{Failure Detection} that uses segmentation-based target identification; (2) \textbf{Failure Case Analysis} through chain-of-thought reasoning to understand the environmental context; (3) \textbf{Movement Suggestions} that provide structured action plans with directional, environmental, and conditional triggers; and (4) \textbf{Memory-Augmented Self-Reflection} that plans recovery sequences and optimizes actions based on stored experiences. (5)\textbf{Reflection Insights Generation} that summarizes the entire recovery plan and gives an adjustment suggestion when the plan fails. The framework enables agents to recover from occlusions and extended target loss by leveraging the reasoning capabilities of VLMs, e.g., GPT-4o, memory management for historical context, self-reflection for continuous improvement, and a robust visual tracking policy for sustained tracking once recovery is achieved.}
    \label{fig:framework}
\end{figure*}

\section{Problem Statement}
We extend the previous Embodied Visual Tracking (EVT) task formulation by modeling it as a Partial Observable Markov Decision Process(POMDP), with state $s \in \mathcal{S}$, partial observation $o \in \mathcal{O}$, action $a \in \mathcal{A}$, and reward $r \in \mathcal{R}$. When the agent receives an observation $o_{t}$, it follows a two-phase strategy to predict the appropriate actions:
\begin{itemize}
    \item \textbf{Tracking Phase}: If the target is visible, the agent uses the visual tracking policy $\pi(a_{t}|s_{t})$ to adjust its actions and maintain a suitable position relative to the target.
    \item \textbf{Recovery Phase}: When the tracking policy corrupts, the agent switches to the recovery phase. During this phase, the agent uses the last three frames of historical observations $(o_{t-2i},o_{t-i},o_t)$, where $i$ is the sampling interval, to predict a recovery action sequence $(a_t, a_{t+1},...a_{t+r})$. This sequence of actions is designed to help the agent recover the target by re-engaging in the search process. If the target reappears, the agent reverts back to the tracking phase and resumes using $\pi(a_{t}|s_{t})$.
\end{itemize}

At each step of the agent’s interaction with the environment, it receives a step reward $r(s_t)$ based on its current position related to target, defined as: $r = 1- \frac{\vert\rho - \rho^*\vert}{\rho_{max}} - \frac{\vert\theta-\theta^*\vert}{\theta_{max}}$, where $(\rho, \theta)$ denotes the current target distance and angle relative to the tracker. $(\rho^*, \theta^*)$ represents the expected spatial position between the target and agent, $(\rho_{max}, \theta_{max})$ represents the agent's visible area. In the tracking phase, the reward encourages the agent to minimize the deviation from the desired target position, while in the recovery phase, the reward encourages the agent to bring back the target into view. The reward function ensures that the agent can successfully balance tracking and recovery behaviors. The ultimate goal of the agent is to maximize the cumulative reward over time by efficiently tracking the target and recovering from failures, i.e., $E_{\pi} \left[\sum_{t=0}^T r_t\right]$.

\section{Embodied Visual Tracking with self-improving Visual Language Models}
In this section, we introduce our proposed hierarchical framework for assisting an embodied visual tracking agent to recover from failure, as illustrated in Figure ~\ref{fig:framework}. When the target can be detected by the vision module, we an off-the-shelf active visual tracking policy to control the movement of the robot, ensuring fast, precise control. When the target disappears from view, the recovery phase is triggered. Our VLM-based reasoning system analyzes the visual scene, interprets environmental cues, and generates a structured recovery strategy by drawing on both its reasoning capabilities and past experiences stored in memory. After executing the recovery action sequence, the VLM performs a critical self-reflection process, evaluating the effectiveness of its decisions for further improvements. These insights are recorded in memory, creating a continuously expanding knowledge base that enables the agent to refine its recovery strategies over time and adapt to novel failure scenarios. This self-improving cycle represents a fundamental shift in EVT approaches—transforming tracking failures from terminal conditions into learning opportunities that progressively enhance the system's robustness in complex, dynamic environments.

\subsection{Visual Tracking Policy}
Our framework employs a modular design where the base visual tracking agent can be substituted with different tracking algorithms. For our primary implementation, we follow the training paradigm from \cite{zhong2024empowering}, which offers robust performance in standard tracking scenarios. This modularity enables us to evaluate different tracking approaches through ablation studies, including traditional approaches like PID controllers, to isolate the impact of our recovery mechanism across varying tracking foundations.

\textbf{Tracking Strategy.}
The fundamental role of our tracking policy is to predict a control action $a_t$ based on the current visual observation $o_t$ that actively adjusts the agent's position and orientation, maintaining the target in the central region of the visual field during the tracking process. The visual tracking policy, represented as $\pi_\theta(a_t|o_t)$, forms the core of our tracking strategy regardless of the specific implementation. In our experiments, we implement this tracking policy using two distinct approaches. Our primary implementation extends from the work in \cite{zhong2024empowering}, employing a CNN-LSTM architecture trained through offline reinforcement learning on a diverse trajectory dataset. This network encodes spatial-temporal information from pre-collected trajectories, and the resulting latent features are fed into an actor-critic network to learn a tracking policy. As an alternative implementation for comparison, we also employ a tuned PID controller to track. It uses the same visual representation as the RL-based agent, processes the target's detected bounding box coordinates, and calculates the error between the current position and an expected position (center of view). This error signal is then fed into the PID algorithm to generate appropriate movement actions that minimize the positional error over time.

Note that, both implementations integrate the same segmentation model, processing raw RGB images to segmentation masks~\cite{cheng2023tracking} that highlight the target in view. This shared visual processing pipeline ensures a fair comparison between the tracking approaches by guaranteeing consistent visual features. The segmentation-based target identification significantly enhances tracking robustness in visually complex environments and provides a reliable foundation for our failure detection.

\textbf{Failure Detection.} 
Our system leverages segmentation-based target identification to implement a straightforward yet effective failure detection mechanism. Specifically, the tracking agent continuously monitors the visibility of the target's segmentation mask within the current field of view. When this mask becomes invisible for more than 3 consecutive steps as the segmentation mask illustrated in Figure ~\ref{fig:framework}, the agent declares a tracking failure and automatically transitions to the recovery phase. We empirically determined this 3-step threshold through extensive testing, finding that the temporal motion patterns learned by the policy network enable it to maintain reasonable movement direction for a short period after target loss. This brief tolerance window prevents premature recovery triggering during momentary occlusions while ensuring timely responses to genuine tracking failures.

\subsection{Vision-Language Reasoning for Failure Recovery} 
When the tracking system detects a failure, we engage in a structured VLM-based reasoning process to maximize recovery effectiveness. Our approach divides the recovery phase into two interconnected modules that leverage the VLM's reasoning capabilities to generate contextually appropriate recovery strategies.

\textbf{Failure Cause Analysis.}
Despite VLMs' powerful reasoning capabilities, directly analyzing a single current observation frame to determine failure causes remains challenging. To address this limitation, we provide the VLM with three continuous observations $\{o_{t-5k}\}_{k=0}^{3}$ sampled at five-step intervals leading up to the failure. These frames are concatenated to create a comprehensive visual context that reveals target movement patterns and environmental changes.

We implement a chain-of-thought reasoning approach by structuring the analysis into three sequential questions, each producing a specific component of our failure analysis:
\begin{enumerate}
    \item Is the target occluded by surrounding environmental structures or objects? → $\beta_{occ} \in \{0,1\}$
    \item If occluded ($\beta_{occ}=1$), describe the occluding object type and its relative position → $E_{obj}$
    \item If not occluded ($\beta_{occ}=0$), describe the target's last known position → $L_{tgt}$
\end{enumerate}

This structured reasoning produces a failure context tuple $\Psi = (\beta_{occ}, E_{obj}, L_{tgt})$, where either $E_{obj}$ or $L_{tgt}$ contains null values depending on the occlusion status $\beta_{occ}$. These context tuples are stored in memory $\mathcal{M}$ for future retrieval when facing similar failure scenarios in the future, enabling recovery improvements.

\textbf{Directional Movement Suggestions.}
Based on the failure context $\Psi$, we prompt VLM to generate a movement plan $\Gamma(\Psi)$ that contains spatially anchored instructions. Each movement plan consists of a sequence of navigational instructions $\Gamma(\Psi) = \{I_1, I_2, ..., I_n\}$ for the $n$ steps. To be specific, the generated suggestion of each step $I_i$ is represented by a tuple $(D_i, Z_i, C_i)$, where $D_i$ is a primary moving direction, e.g., ``move forward", ``turn right", $C_i$ is a conditional trigger, e.g., ``until reaching", ``after passing", and $Z_i$ is an environmental reference object as the landmark, e.g., ``column", ``doorway".

For example, rather than simply suggesting ``move right", the system might recommend $I_1 = \text{\{\texttt{Move forward}, \texttt{Until passing}, \texttt{Right column}\}}$ followed by $I_2 = \text{\{\texttt{Turn right}, \texttt{Null}, \texttt{Null}\}}$. These environmentally anchored instructions provide more actionable guidance by incorporating spatial relationships between the agent's trajectory and observable landmarks, substantially improving the agent's ability to navigate toward the likely target location.

\subsection{Self-Improving with Memory-Augmented Reflection}
\label{Memory-Augmented Reflection}
\textbf{Recovery Action Planning with Memory Retrieval.}
After receive the movement plan $\Gamma$, the VLM generates a concrete recovery action sequence $R = (a_1, a_2, a_3, a_4, a_5)$ where each $a_i \in \mathcal{A}$, $\mathcal{A}$ containing six executable actions:

\begin{itemize}
\item \texttt{Move Forward}: Propel the agent forward by $1 m$;
\item \texttt{Move Backward}: Propel the agent backward by $1 m$;
\item \texttt{Turn Left}: Rotate the agent left by 30 degrees;
\item \texttt{Turn Right}: Rotate the agent right by 30 degrees;
\item \texttt{Jump Over}: Leap over small obstacles within jumping range (if the agent can jump).
\end{itemize}

This fixed-length sequence provides a balance between recovery effectiveness and computational efficiency. To enhance the rationality of the recovery action sequence, we revise the generated $\Gamma$ and $R$ by informing the reflection insights of past similar cases in our memory $\mathcal{M}$. In practice, we compute a similarity score between the current case and historical examples using: $score_{i} = \text{sim}_{text}(\Psi_{cur}, \Psi_i) +\text{sim}_{text}(\Gamma_{cur}, \Gamma_i)$, where $score_{i}$ represents the similarity between the current case and $i_{th}$ experience in memory, $sim_{text}$ are the cosine similarity of two TF-IDF vectors. We empirically retrieves the top-3 most similar cases $\mathcal{N} = \{(\Psi_j, \Gamma_j, R_j, E_j)\}_{j=1}^3$ from memory, where $E_j$ is the reflection insight generated post-recovery. These cases are presented to the VLM as exemplars, enabling the refinement of the recovery action sequence:
 $R_{refined} = f_{VLM}(\Psi_{cur}, \Gamma_{cur}, R_{cur}, \mathcal{N})$

\textbf{Memory Management with Reflection.} Memory management serves as a foundation of our recovery framework, accumulating historical tracking failure cases and recovery experiences that inform future recovery strategies ~\cite{gao2024memorysharinglargelanguage}. It provides a structured repository of past failure-recovery episodes that enables the system to learn from previous successes and failures. Each memory entry $m_i \in \mathcal{M}$ encapsulates a complete recovery episode: $m_i = (\Psi_i, \Gamma_i, R_i, E_i)$, where $\Psi_i$ represents the failure context tuple, $\Gamma_i$ is the movement plan, $R_i$ the executed action sequence, and $E_i$ the reflection insights generated post-recovery, if recovery succeeds, the $E_{i}$ is filled by \texttt{null}.

\textbf{Reflection Insight.} At the end of each recovery attempt, if the recovery fails, the system evaluates the outcome and generates reflection insights. The reflection insights $E_i$ are generated by prompting the VLM to analyze the recovery execution trace $\tau_i = (o_t, a_1, o_{t+1}, a_2, ..., a_5, o_{t+5})$, where $o_t$ represents observations and $a_j \in R_i$ represents actions, resulting in a concise text summarization. $E_i = f_{VLM}(\Psi_i, \Gamma_i, R_i, \tau_i)$.
These insights articulate which actions were effective or ineffective, environmental factors that influenced recovery success, and generalizable principles for similar situations.

To this end, such memory-augmented planning creates a self-improving loop where each recovery attempt contributes to the system's growing expertise in handling complex tracking failures. The memory particularly benefits the handling of long-tail failure cases that standard tracking algorithms struggle to address.

\begin{figure}
    \centering
    \includegraphics[width=0.98\linewidth]{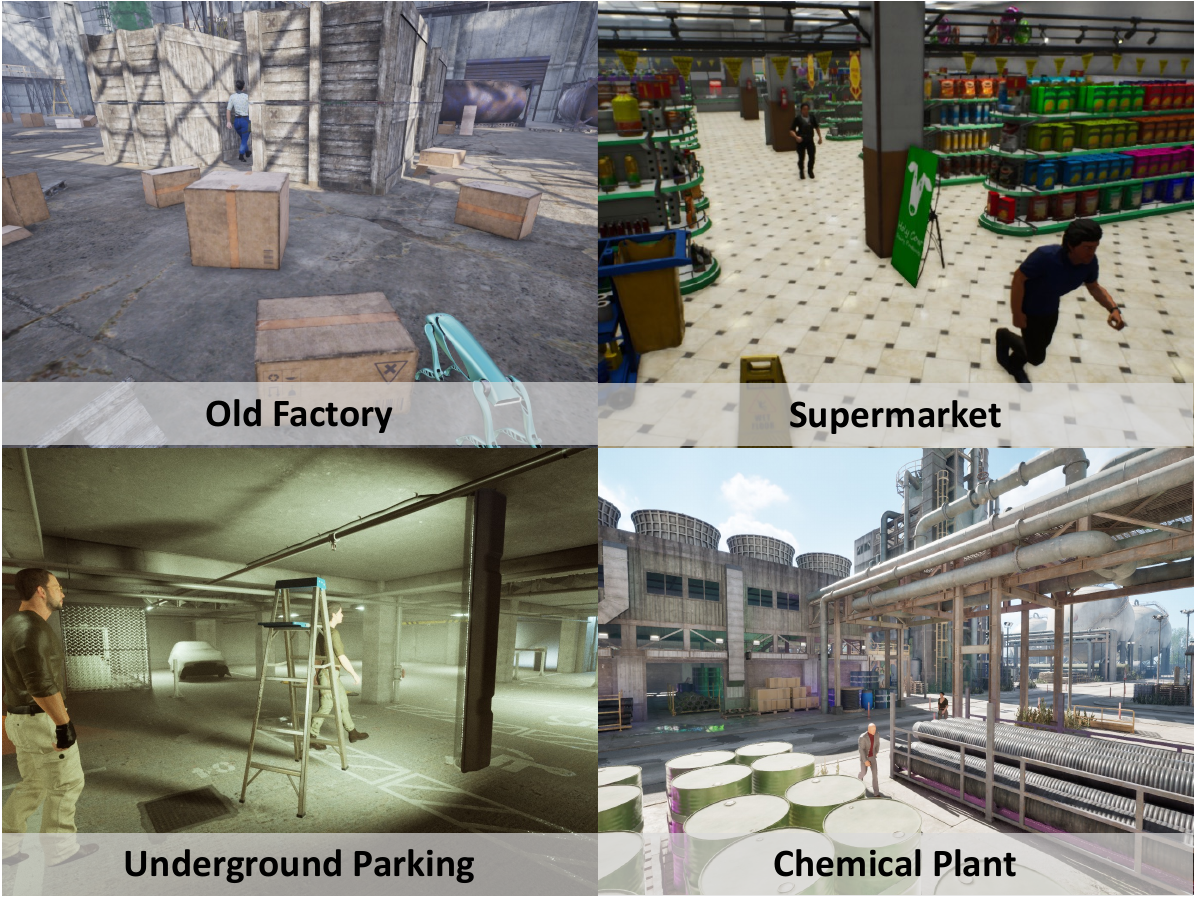}
    \vspace{-0.3cm}
    \caption{Four high-fidelity virtual environments used for testing the embodied visual tracking agents. The four environments are built on Unreal Engine 5 and UnrealCV~\cite{qiu2017unrealcv} to simulate real-world challenges.}
    \label{fig:environments}
    \vspace{-0.4cm}
\end{figure}

\begin{table*}[!t]
    \centering
    \caption{Performance comparison of our VLM-enhanced tracking framework against baselines. Metrics include Average Episodic Reward (ER), Average Episode Length (EL), and Success Rate (SR). The second row for each of our methods shows percentage improvements over corresponding baselines.}
    \begin{tabular}{l|cccc} \hline
         & \texttt{Old Factory} & \texttt{Underground Parking} & \texttt{Supermarket} & \texttt{Chemical Plant} \\ 
       Method & (ER/EL/SR) & (ER/EL/SR) & (ER/EL/SR) & (ER/EL/SR) \\ \hline
        PID & 92/330/0.40 & 87/337/0.38 & 98/253/0.42 & 54/118/0.10 \\ 
        OpenVLA~\cite{kim2024openvla} & -12/162/0.08 & -32/170/0.00 & -9/191/0.06 & -6/198/0.04 \\ 
        GPT-4o & -25/213/0.18 & -16/237/0.30 & 18/225/0.20 & -4/182/0.12 \\ 
        SOTA RL~\cite{zhong2024empowering} & 296/417/0.76 & 270/436/0.68 & 229/408/0.60 & 181/361/0.36 \\ \hline
        
        \textbf{Ours (PID)} & \textbf{214}/\textbf{442}/\textbf{0.74} & \textbf{196}/\textbf{405}/\textbf{0.70} & \textbf{168}/\textbf{404}/\textbf{0.54} & \textbf{108}/\textbf{328}/\textbf{0.32} \\
        \textit{vs. PID} & \textit{+133\%/+34\%/+85\%} & \textit{+125\%/+20\%/+84\%} & \textit{+71\%/+60\%/+29\%} & \textit{+100\%/+178\%/+220\%} \\ \hline

        \textbf{Ours (SOTA RL)} & \textbf{315}/\textbf{484}/\textbf{0.92} & \textbf{313}/\textbf{470}/\textbf{0.88} & \textbf{325}/\textbf{493}/\textbf{0.94} & \textbf{247}/\textbf{403}/\textbf{0.62} \\
        \textit{vs. SOTA RL} & \textit{+6\%/+16\%/+21\%} & \textit{+16\%/+8\%/+29\%} & \textit{+42\%/+21\%/+57\%} & \textit{+37\%/+12\%/+72\%} \\ \hline
        
        Ours w/o Reflection & 259/420/0.68 & 290/449/0.82 & 251/416/0.68 & 255/405/0.58 \\ 
        Ours w/o Memory Retrieval & 298/464/0.78 & 288/462/0.78 & 254/440/0.76 & 198/368/0.44 \\ 
        \hline
    \end{tabular}
    \label{tab:main_result}
\vspace{-0.1cm}
\end{table*}

\section{Experiment}
In this section, we evaluate our framework through comprehensive experiments designed to verify three key claims: 1) Our framework can generally assist off-the-shelf visual tracking to recovery from failure to improve the overall performance, and achieving state-of-the-art results in challenging environments; 2) The key contributed modules in our self-improving framework are effective to the gain of overall tracking performance; 3) The recovery process conducted by our framework is effective in qualitative and quantitative.

\subsection{Experiment Setup}
\textbf{Environment.}
We conduct experiments across four diverse, high-fidelity virtual environments released by UnrealZoo ~\cite{zhong2024unrealzooenrichingphotorealisticvirtual}  to simulate real-world tracking challenges:
\begin{itemize}
    \item \textit{Old Factory}: An abandoned factory scene featuring numerous steel pillars and floor debris. The primary challenge is occlusion from irregular obstacle structures.
    \item \textit{Supermarket}: An indoor retail environment with densely arranged shelving units, challenging tracking with frequent occlusions and similar-looking objects.
    \item \textit{Underground Parking}: A dimly lit parking facility with supporting columns. The main challenges include poor lighting conditions and structural occlusions.
    \item \textit{Chemical Plant}: An expansive chemical facility spanning approximately 4 square kilometers. This environment presents complex challenges including long-distance tracking, indoor-outdoor transitions, and vertical traversal via staircases.
\end{itemize}
These environments, shown as Figure~\ref{fig:environments}, collectively represent the diverse challenges tracking agents encounter in real-world settings and allow us to comprehensively evaluate our framework's robustness and effectiveness.

\textbf{Evaluation Metrics.}
Following established evaluation protocols in embodied visual tracking~\cite{zhong2019ad, zhong2024empowering}, we set the episode length to 500 steps with termination if the target is continuously lost from view for more than 50 steps. The agent's visible area is a fan-shaped area with a radius of $7.5m$ and a maximum viewing angle of 90 degrees. We employ three key metrics: 1)~\texttt{Success Rate (SR)}: The percentage of episodes reaching the maximum 500 steps without failing, measured across 50 episodes. This is our primary metric to reflect the ability to recover from failure. 2)~\texttt{Average Episode Length (EL)}: The average number of steps completed before termination, reflecting sustained tracking capability. 3)~\texttt{Average Episodic Reward (ER)} calculates the average episodic reward over 50 episodes, indicating overall tracking quality.

\textbf{Baseline.}
We compare our approach against four relevant baselines. 1) \texttt{PID}: A classical control approach that calculates the IoU between the detected target bounding box and a desired bounding box position, using a PID controller to adjust agent actions to maximize this IoU. 2) \texttt{OpenVLA}~\cite{kim2024openvla}: While primarily designed for robotic manipulation, we fine-tuned this popular model on our collected dataset to adapt it for tracking tasks, providing a comparison with a recent domain-adjacent approach. 3) \texttt{GPT-4o}: A direct application of the state-of-the-art vision-language model using the same action space as our framework's reasoning module. This baseline demonstrates the raw capabilities of advanced VLMs without our specialized recovery framework. 4) \texttt{SOTA RL}~\cite{zhong2024empowering}: The state-of-the-art RL-based method that integrates the visual foundation model and offline reinforcement learning to learn a generalizable tracking policy. We follow the official repository's data collection and training procedures, using 50k-step trajectories collected in a room with domain randomization for training. 

\textbf{Implementation Detail.} All experiments are conducted on an Nvidia RTX 4090, which handles both rendering the high-fidelity environment and executing the visual tracking policy. The agent takes RGB images as input signals. Its movements are controlled by two parameters: angular velocity and linear velocity. Angular velocity is limited to a range of $-30^{\circ}/s$ and $30^{\circ}/s$, 
while linear velocity ranges from $-1\ m/s$ to $1\ m/s$. During the tracking phase, the PID, OpenVLA, and RL methods operate within this continuous action space, while GPT-4o employs a six-dimensional discrete action space, as detailed in Section ~\ref{Memory-Augmented Reflection}. In the failure recovery phase, the system uses the discrete action space. For failure recovery reasoning, we utilize GPT-4o as the VLM model. We have open-sourced the code and more video capture sequences on the \href{https://sites.google.com/view/evt-recovery-assistant}{project website}.

\begin{figure*}
    \centering
    \includegraphics[width=1.0\linewidth]{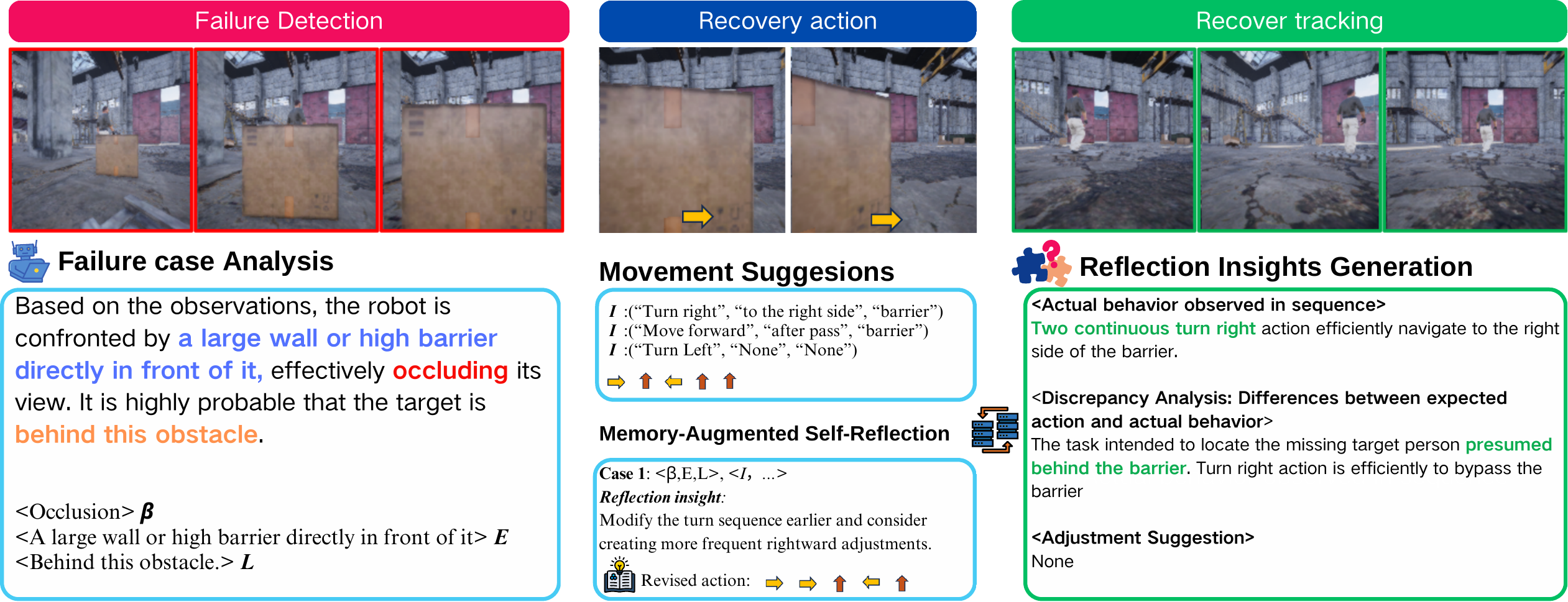}
    \vspace{-0.5cm}
    \caption{Visualization of a successful recovery sequence in \textit{Old Factory}. \textcolor[RGB]{200,50,50}{Red} frames show failure events. \textcolor[RGB]{0,0,168}{Blue}  frames in the middle indicate recovery action (orange arrows). \textcolor[RGB]{0, 168, 0}{Green} frames show successful target reacquisition. The bottom section presents the system's reasoning process, including failure case analysis, movement suggestions, memory-augmented self-reflection, and reflection insights generation that compares expected versus actual behavior. }
    \label{fig:case_analysis}
    \vspace{-0.2cm}
\end{figure*}

\subsection{Main Results}
Table~\ref{tab:main_result} presents a comprehensive comparison between our proposed framework and all baseline methods across four testing environments. The results demonstrate the following key findings:
\textbf{1) Vanilla VLM is not ready for embodied tracking.} OpenVLA and GPT-4o achieved substantially lower performance on the three metrics compared to other methods. Despite using fine-tuned OpenVLA, it still easily lost targets after a period of tracking. The performance of GPT-4o reveals that despite inherent advantages in zero-shot generalization and reasoning capabilities, it exhibits limitations in action prediction accuracy. This limitation leads to error accumulation during extended tracking sequences, ultimately resulting in poor overall performance in long-term tracking tasks. 
\textbf{2) Our VLM-enhanced recovery framework can make the state-of-the-art tracker stronger.} Our implementation with SOTA RL outperformed all baselines, achieving improvements from $21\%$ to $72\%$ against the state-of-the-art RL method~\cite{zhong2024empowering}.
\textbf{3) Our framework can assist different trackers.} We further integrate our framework with the tracker using PID controller, i.e.,\textit{Ours(PID)}.
Our observations indicate that while PID controllers can achieve near-perfect tracking in unobstructed scenes, they rapidly lose targets when faced with sudden direction changes or occlusions. We can see that our framework can significantly improve the performance of the PID tracker, highlighting the potential of combining advanced visual reasoning models with traditional control methods in complex scenes.

\subsection{Ablation Studies}
To evaluate the contribution of key components in our framework, we conducted ablation studies by removing the reflection mechanism and memory retrieval functionality, as shown in the bottom two rows in Table ~\ref{tab:main_result}. Removing the reflection module ``w/o Reflection" resulted in significant performance drops across all environments, with notable decreases in success rates, particularly in the \textit{Old Factory} and \textit{Supermarket}. This indicates that self-reflection on past failures substantially improves the agent's ability to generate effective recovery strategies. When disabling memory retrieval ``w/o Memory Retrieval", we observed substantial performance degradation, especially in the \textit{Chemical Plant}, highlighting that access to past experiences provides crucial contextual information for reasoning. Interestingly, the performance impact suggests that without the explanatory insights from reflection, raw past experiences alone might sometimes mislead the reasoning process in complex scenarios. Notably, both ablated variants still outperformed all baseline methods, confirming that even our reduced framework provides significant advantages over existing approaches. These results highlight the complementary nature of our memory and reflection components, with their combination yielding the strongest overall performance.

\begin{table}[t]
\caption{Recovery performance statistics across four environments. \textbf{Recovery Attempts:} The total number of recovery phases triggered across 50 evaluation episodes.\textbf{Successful Recovery Rate:} The percentage of attempts where the agent successfully brought the target back into view }
    \centering
    \begin{tabular}{c|c|c} \hline 
        Environment & Recovery Attempts& Success Rate \\ \hline
        OldFactory &23&56.5\% \\
        UndergroundParking &28&71.4\% \\
        Supermarket &23&52.2\% \\
        ChemicalPlant &57&57.9\% \\ \hline

    \end{tabular}
    \vspace{-0.4cm}
    \label{tab:quality}
\end{table}

\subsection{Recovery Process Analysis}
To provide deeper insights into our framework's recovery capabilities, we record the total number of recovery phases triggered across 50 evaluation episodes per environment and the percentage of attempts where the agent successfully brought the target back into view. Table ~\ref{tab:quality} presents detailed recovery statistics demonstrating our framework's effectiveness in handling tracking failures. First, our method demonstrates consistent recovery capabilities across all environments, with success rates ranging from 52.2\% to 71.4\%. This is particularly significant considering that baseline methods typically fail to recover once the target is lost from view. The \textit{Underground Parking} shows the highest recovery rate (71.4\%), despite its challenging lighting conditions, suggesting that our VLM reasoning effectively handles visual challenges when structural features remain distinguishable. \textit{Chemical Plant} triggered the most recovery attempts (57), more than double the other environments, highlighting its substantially higher complexity due to its expansive area and diverse structural elements. Despite this complexity, our framework still achieved a 57.9\% recovery rate, demonstrating robust performance even in the most challenging scenarios.

Figure~\ref{fig:case_analysis} illustrates a representative recovery sequence, showcasing our framework's ability to efficiently handle the failure case. The sequence depicts an occlusion scenario in the Old Factory environment, where a large wooden barrier completely blocks the target from view. The visualization is divided into three key phases: Failure Detection (red frames) where the system identifies the occlusion problem and infers the target is behind the obstacle; Recovery Action (blue frames with orange arrows) showing the execution of a "turn right" to navigate around the barrier, and Recovery Tracking (green frames) demonstrating successful target reacquisition. This case analysis reveals how the robot, through an effective reasoning process, determines the direction of movement and the optimal action sequence to overcome the barrier.

\textbf{Failure Cases.} We observe that unsuccessful recovery predominantly occurred in cases with multiple simultaneous occlusions, where the VLM reasoning cannot correctly identify which occlusion the target moved behind. These findings indicate promising directions for future work on more sophisticated environmental modeling that can better understand complex spatial relationships. Future implementations might also benefit from extending the contextual memory to include environmental maps built during tracking, allowing for more informed navigation decisions during recovery phases.

\section{Conclusion}
In this paper, we presented a novel VLM-enhanced framework for robust embodied visual tracking in complex environments. Our approach successfully addresses a critical limitation in existing tracking systems by integrating vision-language reasoning capabilities with a memory-based reflection mechanism to recover from tracking failures. The experimental results across diverse environments demonstrate our method's superior performance compared to state-of-the-art approaches, particularly in challenging scenarios involving occlusions and sudden target disappearance. While our method improves the tracking success rate, cases with multiple simultaneous occlusions remain challenging. These insights open promising avenues for future work on refining VLM-driven perception and recovery mechanisms.

Though our work has made considerable progress in embodied visual tracking, significant challenges remain. For example, while Vision-Language Models (VLMs) have demonstrated impressive cross-domain generalization and reasoning capabilities, their computational demands present obstacles for efficiently tracking in high-dynamic scenarios. So we will further explore methods to speed up the reasoning process for robots. 
Moreover, we will further adapt such a self-improving framework for other tasks, such as navigation and human-robot interaction tasks, suggesting broader applicability of our framework across embodied AI challenges.

\addtolength{\textheight}{-12cm}   







\bibliographystyle{plain}
\bibliography{ref}

\end{document}